\documentclass[11pt,a4paper]{article}
\usepackage[hyperref]{ranlp2021}
\usepackage{times}
\usepackage{latexsym}
\usepackage{multirow}
\usepackage{multicol}
\usepackage{tabularx}
\usepackage{graphicx}
\usepackage{booktabs}
\usepackage{mathtools}

\usepackage{enumitem}

\usepackage{microtype}

\aclfinalcopy % Uncomment this line for the final submission
\def\aclpaperid{175} %  Enter the acl Paper ID here

\title{Predicting the Factuality of Reporting of News Media\\
Using Observations About User Attention in Their YouTube Channels}

\author{

Krasimira Bozhanova \\
FMI, Sofia University \\
``St. Kliment Ohridski'' \\
Sofia, Bulgaria\\
{\small \texttt{krasimira.bozhanova@gmail.com}}
  \And
  Yoan Dinkov \\
  FMI, Sofia University \\ ``St. Kliment Ohridski'' \\
  Sofia, Bulgaria\\
{\small \texttt{yoandinkov@gmail.com}}
 \And
  Ivan Koychev \\
  FMI and GATE, Sofia University \\``St. Kliment Ohridski'' \\
  Sofia, Bulgaria\\
 {\small \texttt{koychev@fmi.uni-sofia.bg}}
 \AND
  Maria Castaldo \\
  Univ. Grenoble Alpes, \\ CNRS, Inria,  Grenoble INP, \\ GIPSA-lab, F-38000, France \\
  {\small \texttt{maria.castaldo@grenoble-inp.fr}}
  \\\And
  Tommaso Venturini \\
  Univ. Grenoble Alpes, \\ CNRS, Inria, Grenoble INP, \\ GIPSA-lab, F-38000, France \\
  {\small \texttt{tommaso.venturini@cnrs.fr}}
  \\\And
  Preslav Nakov \\
  Qatar Computing Research Institute\\ HBKU\\
  Doha, Qatar\\
  {\small \texttt{pnakov@hbku.edu.qa}}
}

\date{}

\begin{document}
\maketitle
\begin{abstract}
We propose a novel framework for predicting the factuality of reporting of news media outlets by studying the user attention cycles in their YouTube channels. In particular, we design a rich set of features derived from the temporal evolution of the number of views, likes, dislikes, and comments for a video, which we then aggregate to the channel level. We develop and release a dataset for the task, containing observations of user attention on YouTube channels for 489 news media. Our experiments demonstrate both complementarity and sizable improvements over state-of-the-art textual representations. 
\end{abstract}

\section{Introduction}
\label{sec:introduction}

Disinformation in the news and in social media is perceived as having a major impact on society, e.g.,~during the 2016 US Presidential election~\cite{Grinberg374} and the Brexit referendum~\cite{10.1007/978-3-030-01129-1_17}. During the COVID-19 pandemic outbreak, the moral panic~\cite{mcluhan1964understanding} around online disinformation grew to a whole new level as \emph{the first global infodemic}.\footnote{MIT Technology Review: \url{tinyurl.com/y8oschng}} Nowadays, fighting disinformation online has been recognized as one of the most important issues societies around the world are facing today.

In this paper, we highlight an aspect of disinformation that is often neglected. Rather than examining the truth-value of individual piece of information, we investigate the general quality of the attention regimes in different news outlets, by analyzing news media's YouTube channels.

YouTube is the largest and the most popular platform for video sharing with over two billion users and it is also the second most widely used news source in USA, after Facebook.\footnote{http://tinyurl.com/y4apu58j} While the platform has been scrutinized for the way in which it may amplify marginal and sometimes radical contents \cite{Munn2019,Ribeiro2020}, the connection between attention dynamics and disinformation levels is still largely unexplored.

Here, we do not focus on specific videos, but rather on entire YouTube channels of news media and their attention dynamics. In particular, we are interested in differentiating the ``attention cycles'' \cite{Downs1972,Leskovec2009} of YouTube channels, that is to assess the rapidity and the steepness with which their videos rise and fall in the consideration of their audiences. While some outlets encourage extensive and diverse discussions, other tend to concentrate everyone's attention on the latest hot-button, thus distracting the public opinion instead of nourishing it~\cite{Venturini2019}.

Our contributions are the following:
\begin{itemize}
    \item We propose to model the factuality of news media based on the user attention cycles in their respective YouTube channels.
    \item We release a specialized dataset for the task.\footnote{http://github.com/krasimira-bozhanova/youtube-attention-cycles-dataset}
    \item We show experimentally that considering attention cycles yields considerable performance gains on top of text representations for predicting the factuality of news media.
\end{itemize}

The paper is organized as follows: Section~\ref{sec:related_work} presents related work.  Section~\ref{sec:dataset} describes our dataset. Section~\ref{sec:method} discusses our methodology. Section~\ref{sec:experiments_results} presents the experiments and results. Section~\ref{sec:discussion} offers analysis and discussion. Section~\ref{sec:conclusion} concludes and points to directions for future work.
\section{Related Work}
\label{sec:related_work}

Significant efforts have been dedicated in the last years to automating the detection of \emph{disinformation} (which is commonly referred to as \emph{fake news}), which we look at more closely in this section. At the end, we mention previous attempts to use data from YouTube for media classification tasks.

\paragraph{Analysis of the Content} \label{related-work-content}
Many approaches have been proposed to analyze both the style and the content of fake news. Using natural language processing techniques, \citet{horne_adali_2017} pointed out how fake news can be characterized by stylistic features such as the overuse of proper nouns, punctuation, capital letters, negation terms, and repetitions in the text \cite{rubin_conroy_2016}. 
Fake news has also been associated with intensity of sentiment and emotions, compared to mainstream news
\cite{Crestani:2012}.
Here, we also use textual representations, but (\emph{i})~we focus mainly on analyzing the user attention cycles in YouTube channels, and (\emph{ii})~we aim at categorizing entire news media outlets rather than individual pieces of news.

\paragraph{Analysis of the Response}
Many researchers used the user reactions in social media platforms to identify disinformation, e.g.,~the content and the number of replies to a piece of news, or the propagation of the content in the network. For instance, \citet{zhao_enquiring_2015} classified disputed claims based on their comments and reactions, assuming that, if a claim is not true, at least some replies would question its factuality.
Indeed, later studies \cite{csi,fang} have shown that user response features are quite important.
Here, we also focus on inspecting the user response and its potential link to disinformation, but we use user attention cycles in YouTube.

\paragraph{Analysis of Entire News Media Outlets} \label{media_classifications}

Looking at a news media outlet as a source of low-quality content is another way to approach the problem. In these methods, features modelling the overall trustworthiness of the source are used, such as
(\emph{i})~Does the news media outlet use verified accounts on established platforms, such as Wikipedia and Twitter?

\noindent (\emph{ii})~If it does, do these accounts have a proper description, location, website references, etc? 
(\emph{iii})~How does the URL of the media's website look like?
(\emph{iv})~Does the medium express political bias or sentiment?
\citet{ramy_karadzhov_2018} and \citet{ramy_karadzhov_2020} used features motivated by these questions to achieve better results in combination with content features and user profiles in social media. In our work, we also aim at classifying entire news media outlets, but we do so using user attention cycles in YouTube along with text.

\paragraph{Using Temporal Attention Data for Disinformation Detection}
\label{temporal_data_related_word}
We focus on the analysis of temporal patterns associated with news outlets of different type and quality. Previous studies  \cite{csi,fang} have suggested that a combination of temporal, content-based, and user-based features is promising for disinformation detection. As the dynamics of the viral spread is often associated with successful junk news, we look at studies focusing on modelling virality, such as \cite{twitter-virality-modelling} for tweets. Similar features are well-suited for our task, as described in Section~\ref{sec:attention-features}, but (\emph{i})~we model the user behavior differently, and (\emph{ii})~we focus on data collected from the YouTube channels of the target news media.

\paragraph{Using YouTube Data for Classification}
The YouTube platform contains information that is still underexplored for the purposes of disinformation detection. \citet{youtube-acoustic} looked into detecting the left-centre-right political bias of YouTube channels. \citet{ramy_karadzhov_2020} included features from the news source's YouTube channel, derived from both sound and user profiles. The above work uses raw statistics about the number of views, likes, dislikes, and comments per video. We, instead, use much richer temporality features in combination with the textual representation of the videos.

\section{Data}
\label{sec:dataset}

We started from a corpus of news media outlets, whose reliability has been evaluated by Media Bias/Fact Check\footnote{http://mediabiasfactcheck.com} (MBFC). Lead by a team of independent journalists and researchers, MBFC has analyzed close to 4,000 news outlets over the past six years. For each news outlet, they provide a detailed analysis summarized by a `factuality' score chosen among: \emph{Very High}, \emph{High}, \emph{Mostly Factual}, \emph{Mixed}, \emph{Low}, and \emph{Very Low}. 

We searched the YouTube channels of news outlets in the MBFC corpus and we monitored all the videos they published from February'2020 to August'2020. Using the YouTube Data API,\footnote{http://developers.google.com/youtube/v3/docs/videos} we
collected the number of views, likes, dislikes and comments collected during the first seven days after the publication of each video. We also stored its title and its description.

We observed that the percentage of media channels labelled with the \emph{Very High} and the \emph{Very Low} categories was 3.1\% and 1\%, respectively. We thus merged \emph{Very High} with \emph{High}; \emph{Mostly Factual} with \emph{Mixed}; and \emph{Very Low} with \emph{Low}, ending up with a 3-way labelling: \emph{High}, \emph{Mixed}, and \emph{Low}.

\begin{table}[t]
    \centering
    \begin{tabular}{lrr@{ }@{ }@{ }@{ }@{ }@{ }@{ }@{ }@{ }@{ }rr}
        \toprule
         {\bf Factuality} &
         \multicolumn{2}{c}{\bf Channels} & 
         \multicolumn{2}{c}{\bf Videos} \\
          & \# & \% & \# & \% \\
         \midrule
        High & 308 & 63.0 & 22,932 & 61.7 \\
        Mixed & 153 & 31.3 & 12,125 & 32.7 \\
        Low & 28 & 5.7 & 2,091 & 5.6 \\
        \midrule
        Total & 489 & 100.0 & 37,148 & 100.0 \\
        \bottomrule
    \end{tabular}
    \caption{Statistics about the dataset, showing the distribution of the channels and of the videos for each level of factuality of reporting.}
    \label{table:data-cleaning-1}
\end{table}

Finally, for the sake of data balancing, we excluded the channels with fewer than 20 videos, and we capped the most prolific channels at the newest 100 videos. The final distribution of channels and their corresponding videos for each level of factuality is shown in Table~\ref{table:data-cleaning-1}.

\section{Method}
\label{sec:method}

Our system is composed of two main components focusing on (\emph{i})~data preparation, and on (\emph{ii})~sequential classification, respectively. Below, we describe the representation we use for video-level and also for channel-level classification.

\subsection{Representation}
\label{sec:feature-engineering}

The data preparation component transforms the YouTube source data and produces representations (or features) for our model. We generate representations both for the textual content of the videos and for the user attention data, for which we introduce a number of novel features, presented in Section~\ref{sec:attention-features}.

\subsubsection{Textual Features}
\label{sec:text-features}

We gathered the title and the description of each video. We extracted Sentence BERT embeddings (768 features) for each title and description. These embeddings are derived from a modification of the pretrained BERT model, which yields semantically meaningful sentence embeddings of size 768, which are trained to be readily comparable using cosine similarity \cite{sentence-bert}.

\subsubsection{Attention Features}
\label{sec:attention-features}

We hypothesize that the attention received over time by the videos in a YouTube channel can be used to predict the quality of its contribution to the online public debate, as captured (albeit imprecisely) by the factuality score assigned by MBFC.
We generate a set of features that model the user attention cycles by looking at the temporal variation of the number of \emph{user actions} (Views, Likes, Dislikes, and Comments) in the first week after a video has been published. We aim to model the following:
\begin{itemize}
    \item How are user actions distributed hourly/daily?
    \item How much are the user actions concentrated in peak hours?
    \item At what moment in time does the peak hour for each user action type occur?
    \item How steep is the time series in terms of the distribution of hourly user actions?
\end{itemize}

We define \(UA_{d_i}\) as the total number of user actions that occurred by the end of day \emph{i}. In our case, \emph{i} ranges in $\{1,2,\ldots,7\}$. Similarly, \(UA_{h_j}\) is the number of user actions occurring by the end of hour \emph{j} after the publication of the video. We generate a set of attention features for each user action, which we group into the following categories:
\begin{enumerate}
    \item \textbf{User actions daily percentage} ($D_{d_i}$), or the fraction of user actions out of the total that occurred on day $i$, where $1 \leq i \leq 7$, and $UA_{d_0} = 0$ (7 features per user action):
            \[D_{d_i} = \frac{UA_{d_i} - UA_{d_{i-1}}}{UA_{d_7}}\]
    \item \textbf{User actions daily cumulative percentage} ($DC_{d_i}$), or the fraction of user actions out of the total by the end of day $i$, where $1 \leq i \leq 7$ (7 features per user action):
            \[DC_{d_i} = \frac{UA_{d_i}}{UA_{d_7}}\]
    \item \textbf{User actions daily increase} ($DI_{d_i}$), or the proportion of increase in the number of user actions on day $i$ compared to day $i-1$, where $2 \leq i \leq 7$ (6 features per user action):
            \[DI_{d_i} = \frac{UA_{d_i} - UA_{d_{i-1}}}{UA_{d_{i-1}}}\]
    \item \textbf{User actions hourly increase} ($HI_{h_j}$), or the proportion of increase in the number of user actions during hour $j$ compared to those during hour $j-1$, where $2 \leq j \leq 168$ (167 features per user action):
            \[HI_{h_j} = \frac{UA_{h_j} - UA_{h_{j-1}}}{UA_{h_{j-1}}}\]
    \item \textbf{User actions average hourly increase per day} ($AHI_{d_i}$), or the average hourly increase in the number of user actions on day $i$, where $1 \leq i \leq 7$ (7 features per user action):
            \[AHI_{d_i} = \frac{\sum_{j={(i-1)*24 + 1}}^{n=i*24} HI_{h_j}}{24}\]
    \item \textbf{User actions majority interval length} ($MI_T$), or the number of hours containing the majority of the user actions
    \begin{equation*}
      \begin{multlined}
            MI_T = \min_{1 \leq i < j \leq 168} \left\{ j-i \middle| \frac{UA_{h_j} - UA_{h_i}}{UA_{h_{168}}} \geq T \right\} 
      \end{multlined}
    \end{equation*}
    where $T$ (one of \{0.5, 0.7, 0.9\}) is the \emph{majority} share (3 features per user action).
    \item \textbf{User actions peak delay interval} ($PDI$), or the number of hours leading to the hour with the highest concentration of user actions (1~feature per user action):
            \[PDI = \mathop{\mathrm{argmax}}_{\left\{i|2 \leq i \leq 168\right\}} UA_{h_i} - UA_{h_{i-1}}\]
    \item \textbf{User actions alive interval length} ($AI$), or the hour up to which user actions were recorded (1~feature per user action):
    \begin{equation*}
      \begin{multlined}
            AI = \min_{1 \leq p \leq 167}(p |  UA_{h_i} - UA_{h_{i-1}} = 0, \\
            \forall i \in \{ p+1,\dots,168 \})
      \end{multlined}
    \end{equation*}
    \item \textbf{User actions peak share} ($PS$), or the number of user actions during the peak hour divided by the total (1 feature per user action):
            \[PS = \max_{2 \leq i \leq 168} \left\{ \frac{UA_{h_i} - UA_{h_{i-1}}}{UA_{h_{168}}} \right\} \]
\end{enumerate}

\begin{figure}[t]
    \centering
    \includegraphics[width=8cm]{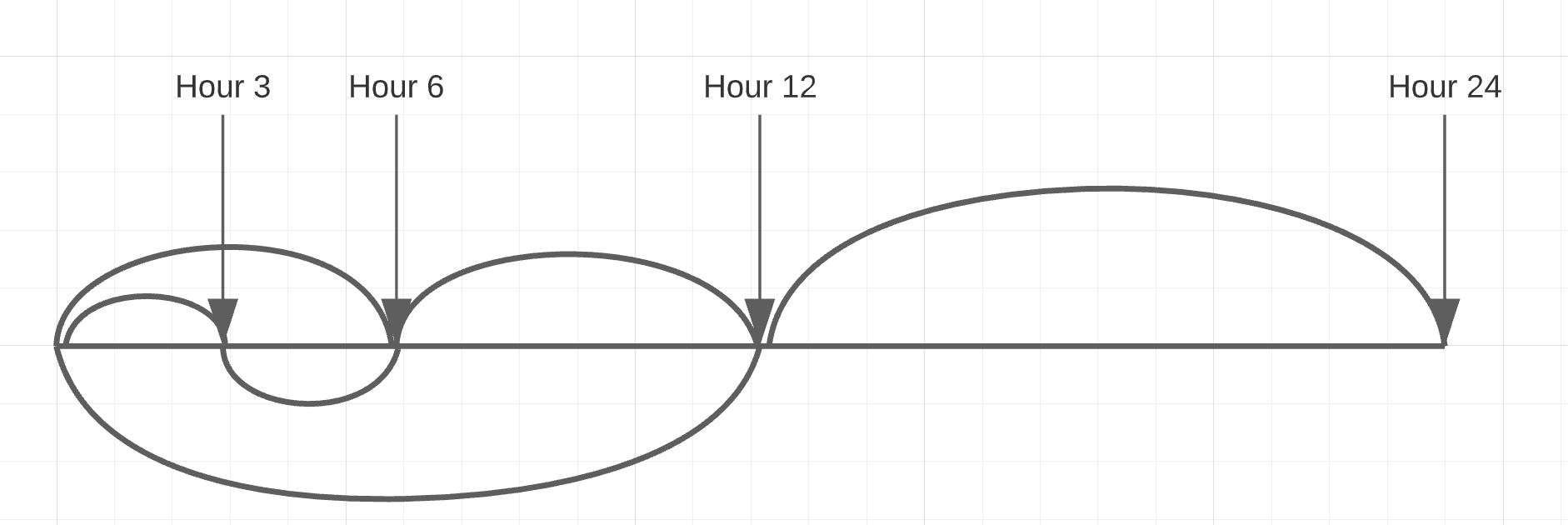}
    \caption{First day breakdown into periods}
    \label{fig:first-day-periods}
\end{figure}

Most of the attention received by the videos in our corpus is concentrated in the first day after a video has been published, and thus we monitor the attention during this period more closely. Besides \emph{daily} and \emph{hourly}, we look at six additional periods during the first day, as depicted on Figure~\ref{fig:first-day-periods}. We extract the following features: (\emph{i})~\emph{Percentage of User Actions per Period}, (\emph{ii})~\emph{User Actions per Period Increase}, and (\emph{iii})~\emph{User Actions Average Hourly Increase for a Period}. This yields 18 additional features and a total of 218 features per user action.

To model the opinion of the users regarding the videos, we also use features derived by ratios of different user action types:
\begin{itemize}
    \item \emph{Positive Reactions}: the ratio between the number of likes and the number of views;
    \item \emph{Negative Reactions}: the ratio between the number of dislikes and the number of views;
    \item \emph{Engagement}: the ratio between the number of comments and the number of views;
    \item \emph{Controversiality}: the ratio between the number of likes and the sum of the number of likes and of dislikes.
\end{itemize}

For each of these ratios, we calculate a set of features that show how the numbers change daily, similarly to the \emph{User Actions Daily Percentages} (7~features per ratio) and the \emph{User Actions Daily Cumulative Percentages} (6~features per ratio) features. We further generate ratio features for the more granular first day periods (6~features per ratio), which yields a total of 19 ratio-driven features. 
Overall, we have 952 attention features per video.

\subsection{Models}
\label{sec:models}

Our architecture contains two consecutive classifications: (\emph{i})~for YouTube videos, and (\emph{ii})~for YouTube channels. As we want to make use of the features derived from the YouTube videos, we labelled each video with the factuality score of the channel that published it, using distant supervision. 

Thus, the video classification learns to predict the factuality labels that are projected from the corresponding channels. Naturally, not all videos published by a low-factuality channel necessarily contain disinformation. Yet, this is not a problem since we do not aim at classifying correctly individual videos, but at detecting factuality-related patterns, which would then be used at the channel level: our channel classifier uses the predictions of the video-level classifier to predict the factuality of channels.

\subsubsection{Video Modelling}
\label{sec:video-features}
For each video, we have 768 features from the sentence-level BERT representation.
We calculated these features once  for the title and once for the description of the video, obtaining a total of 1,536 textual features.

We further have 952 attention-driven features per video. To validate the relevance of these features with respect to our classification task, we apply a set of feature selection methods over the training split of our dataset, namely ANOVA, Pearson correlation, and Spearman correlation. According to these methods, the ratio features turn out to be the most relevant ones. We selected the best 100 features from each method or 124 attention video-level features, which we used in our classification experiments. Combined with the 1,536 textual features, this yielded a total of 1,660 features per video.

\subsubsection{Channel Modelling}
\label{sec:channel-features}

For the second classifier, at the channel level, we generate the following groups of features:
\begin{enumerate}
    \item YouTube statistics (total of 13 features):
    \begin{itemize}
        \item Popularity (7 features): number of subscribers, average number of hourly, daily, and weekly views and comments;
        \item Activity (5 features): number of videos, average number of videos published hourly, daily, and weekly, and average number of videos per channel subscriber;
        \item Attention concentration (1 feature): Gini index measuring the concentration of video views within a channel.
    \end{itemize} 
    \item Averaged videos features (1,660) features): average values of the features for the videos published by the channel. For each \emph{video feature}, we have a corresponding aggregated \emph{channel feature}.
    \item Aggregated video-level classifier predictions features (9 features): for each channel, we aggregate the predictions of the video-level classifier for the videos in that channel. We use three types of aggregations:
    \begin{itemize}
        \item maximum probability across the videos for each factuality label;
        \item average probability  across the videos for each factuality label;
        \item factuality distributions percents: for each factuality, this is the percent of videos predicted to have that factuality.
    \end{itemize}
\end{enumerate}

\section{Experiments and Evaluation}
\label{sec:experiments_results}

We train two subsequent classifiers for factuality prediction: for videos and for channels. We conduct experiments with different models and we compare them to a majority-class baseline. We evaluate the models in terms of accuracy, balanced accuracy, and mean absolute error (MAE). MAE is a more relevant measure in our case as it takes into account the ordering of the labels: confusing high factuality with mixed factuality is a smaller error than confusing it with low factuality.

While our ultimate goal is to classify channels, we start with video classification, then we aggregate the predictions, and we use them to make predictions at the channel level.

We divide the dataset into training, development, and test split at the channel level. Then, for the video-level experiments, we  use for training/development/testing the videos for the respective channels. Note that this guarantees that all videos for a given channel go into the same split.

\subsection{Video-Level Classification} 
\label{videos-classification-results}

Below, we report results using Gradient Boosted Decision Trees (GBDT). We also experimented with logistic regression, ordinal logistic regression, and SVM with various kernels, but they performed worse.
We trained separate models (a)~using the textual representation, and (b)~using the user attention cycles. Table~\ref{tab:video-classification-results} shows the evaluation results.

\begin{table}[h]
    \centering
    \begin{tabular}{@{}c@{ }@{ }@{ }lrrr@{}}
         \toprule
         \bf \# & \bf Experiment & \bf Acc. & \bf Bal. Acc. & \bf MAE\\
         \midrule
        0 & Baseline & 61.72 & 33.33 & 0.4391 \\
        1 & BERT & \bf 67.54 & 45.89 & \bf 0.3692 \\
        2 & User attention & 64.93 & \bf 55.12 & 0.3979 \\
        \bottomrule
    \end{tabular}
    \caption{Video-level experiments with GBDT.}
    \label{tab:video-classification-results}
\end{table}

\begin{table*}[t]
    \centering
    \begin{tabular}{lllrrr}
         \toprule
         \bf Group & \bf \# & \bf Experiment & \bf Dim. & \bf Acc. &  \bf MAE\\
         \midrule
        Baseline & 0 & Majority class & - & 63.08 & 0.4308 \\
        \midrule
        \multirow{3}{*}{\it Text} &
        1 & BERT averaged & 1,536 & 73.85 & 0.3077 \\
        & 2 & BERT aggregated predictions & 9 & 75.38 & 0.3077 \\
        \cmidrule{2-6}
        & 3 & BERT all & 1,545 & 73.85 & 0.3538 \\
        \midrule
        \multirow{4}{*}{\it User Attention} &
        4 & User attention averaged &  124 & 75.38 & 0.2769 \\
        & 5 & User attention channel statistics & 13 & 63.08 & 0.4000 \\
        & 6 & User attention aggregated predictions & 9 & 67.69 & 0.3846 \\
        \cmidrule{2-6}
        & 7 & User attention all & 146 & 70.77 & 0.3231 \\
        \midrule
        \multirow{1}{*}{\it Ensemble} &
        8 & BERT all + User attention averaged & 1,669 & \textbf{76.92} & \textbf{0.2615} \\
        \bottomrule
    \end{tabular}
    \caption{Channel-level experiments.}
    \label{table:channel-classification-results}
\end{table*}

Note that our datasets are not well-balanced and have very few examples of low-factuality videos and channels. To mitigate this, we apply over-sampling using SMOTE \cite{smote}, which generates additional synthetic examples. Moreover, it is important that the video classifier generates predictions for the low-factuality and the mixed-factuality classes; otherwise, the predictions for these classes could be lost when aggregating for the channel classification. Thus, we also report \emph{balanced accuracy}, as it is important when choosing which video experiments to select for aggregation.

\subsection{Channel-Level Classification} \label{channel-classification-results}

We experimented with several approaches for channel classification:
\begin{itemize}
    \item using aggregated video-level features to obtain channel-level representation;
    \item using the posterior probabilities of video-level classifiers;
    \item using the previous two together;
    \item ensemble of different channel-level classifiers.
\end{itemize}

For the ensemble aggregation, we experimented with three methods for choosing the most likely class for a given channel:
\begin{itemize}
    \item after averaging the predictions from the various models (mean);
    \item after getting the maximum probability prediction from the various models (max);
    \item after getting the minimum probability prediction from the various models (min) -- tells us which class is least likely to be wrong.
\end{itemize}

The results are shown in Table~\ref{table:channel-classification-results}. All experiments use GBDT, except for experiment~8, which uses ordinal logistic regression. 

\section{Discussion}
\label{sec:discussion}

Below, we analyze the results and we perform an ablation study.

\subsection{Result Analysis}

We can see in Table~\ref{table:channel-classification-results} that all models improve over the majority class baseline by a sizable margin. We further see that using average information on user attention cycles (line~4) performs better than using textual features (lines~1--3). Moreover, combining the two yields the best result (line 8). 
The other two sets of user attention features: channel statistics and aggregated video-classifier predictions, do not contribute to the combined user attentions model (compare line~4 to line~6).

The relative improvements over the majority class baseline in terms of accuracy are generally smaller than those for MAE, which can be explained by class imbalance. To improve accuracy, the models need to learn to assign \emph{mixed} and \emph{low} factuality labels properly (as the majority class is \emph{high} factuality). Most of the trained models undervalue the unrepresented classes. If some of the balancing techniques are applied, the models recognize better the \emph{low} and the \emph{mixed} examples, but at the cost of false positives for these classes from the \emph{high}-factuality examples, which decreases the overall accuracy. In contrast, MAE rewards models that can improve the small classes even given the risk of introducing some errors for the majority class.

\begin{table}[!htb]
    \centering
    \begin{tabular}{lrrr}
    \toprule
    \bf{Feature Group} & \bf{Acc.} & \bf{Bal. Acc.} & \bf{MAE} \\
    \midrule
    Baseline & 63.08 & 33.33 & 0.4308 \\
        \midrule
        Views (V) & 66.15 & 41.79  & 0.3692 \\
        Dislikes (D) & 64.62 & 39.27 & 0.4000 \\
        Comments (C) & 56.92 & 32.64 & 0.4615 \\
        Likes (L) & 64.62 & 37.56 & 0.3846 \\
        \midrule
        V + L + C & 67.69 & 49.27 & 0.3385 \\
        V + L + D + C & 69.23 & 44.27 & 0.3385 \\
        \midrule
        Engagement & 63.08 & 44.27 & 0.4000 \\
        Controversiality & \textbf{70.77} & 48.33 & \textbf{0.3231} \\
        Positive reactions & \textbf{70.77} & 48.33 & \textbf{0.3231} \\
        \midrule
        Contr + Eng & 63.08 & 45.12 & 0.4000 \\
        Contr + Pos & 67.69 & 46.71 & 0.3538 \\
        Pos + Eng & 64.62 & 46.79 & 0.3846 \\
        \midrule
        Channel statistics & 63.08 & 39.31 & 0.400 \\ 
        \midrule
        Aggregated & 67.69 & \textbf{60.20} & 0.3846 \\ 
        \midrule
        All & \textbf{70.77} & 56.02 & \textbf{0.3231} \\
    \bottomrule
    \end{tabular}
        \caption{Ablation study (channel-level classification) using various user attention features.}
        \label{table:ablation-study-channels}
\end{table}

\subsection{Ablation Study}

As our focus is on attention cycles, we performed an ablation study for these features against the combined user attention channel model. The results are shown in Table~\ref{table:ablation-study-channels}. 
We can see that ratio features such as \emph{controversiality} and \emph{positive reactions} alone yield the best accuracy and MAE. Using the predictions of the video-level classifier as features yields the best balanced accuracy. This confirms the importance of having accurate \emph{low}- and \emph{mixed}-factuality predictions for the video classifier prior to the aggregation. Finally, the overall best results, when considering all measures, are achieved when combining all features.

\section{Conclusion and Future Work}
\label{sec:conclusion}

We proposed a novel framework for predicting the factuality of reporting of news outlets by studying the user attention cycles in their respective YouTube channels. We further designed a rich set of features derived from the temporal evolution of the number of views, likes, dislikes, and comments for a video, which we then aggregated at the channel level. Our experiments demonstrated both complementarity and sizable improvements over state-of-the-art textual representations.

We further developed and released a dataset containing observations about user attention on YouTube channels for 489 news media.
We hope that this will enable future research on using data from video sharing platforms.

In future work, we plan to improve the class imbalance of the dataset by extending it with more examples. We further want to integrate additional features based on the comments for the videos and other information sources such as Twitter and Wikipedia. Finally, we plan to study the utility of user attention cycles for other related tasks such as political ideology detection for news media.

\section*{Ethics and Broader Impact}

\paragraph{Data Collection}
Our dataset was collected from YouTube, using their public API.

\paragraph{User Privacy}

Our dataset contains aggregated attention statistics without any user data.

\paragraph{Biases}

Any biases found in the dataset are unintentional, and we do not intend to do harm to any group or individual.

\paragraph{Intended Use and Misuse Potential}

Our dataset and the proposed model can enable the development of systems for automatic detection of reliable/unreliable YouTube channels, which could support media literacy, as well as analysis and decision making for the public good. However, they could also be misused by malicious actors.

\paragraph{Environmental Impact.}

Finally, we would also like to warn that the use of large-scale Transformers requires a lot of computations and the use of GPUs/TPUs for training, which contributes to global warming \cite{strubell-etal-2019-energy}.

\section*{Acknowledgments}

This research is part of the Tanbih mega-project,\footnote{\url{http://tanbih.qcri.org}} developed at the Qatar Computing Research Institute, HBKU, which aims to limit the impact of ``fake news'', propaganda, and media bias by making users aware of what they are reading, thus promoting media literacy and critical thinking.

This research is also partially supported by Project UNITe BG05M2OP001-1.001-0004 funded by the OP ``Science and Education for Smart Growth'' and co-funded by the EU through the ESI Funds.

\bibliographystyle{acl_natbib}
\bibliography{bib/main}

\begin{thebibliography}{21}
\expandafter\ifx\csname natexlab\endcsname\relax\def\natexlab#1{#1}\fi

\bibitem[{Baly et~al.(2018)Baly, Karadzhov, Alexandrov, Glass, and
  Nakov}]{ramy_karadzhov_2018}
Ramy Baly, Georgi Karadzhov, Dimitar Alexandrov, James Glass, and Preslav
  Nakov. 2018.
\newblock Predicting factuality of reporting and bias of news media sources.
\newblock In \emph{Proceedings of the 2018 Conference on Empirical Methods in
  Natural Language Processing}, EMNLP~'18, pages 3528--3539, Brussels, Belgium.

\bibitem[{Baly et~al.(2020)Baly, Karadzhov, An, Kwak, Dinkov, Ali, Glass, and
  Nakov}]{ramy_karadzhov_2020}
Ramy Baly, Georgi Karadzhov, Jisun An, Haewoon Kwak, Yoan Dinkov, Ahmed Ali,
  James Glass, and Preslav Nakov. 2020.
\newblock What was written vs. who read it: News media profiling using text
  analysis and social media context.
\newblock In \emph{Proceedings of the 58th Annual Meeting of the Association
  for Computational Linguistics}, ACL~'20, pages 3364--3374.

\bibitem[{Chawla et~al.(2002)Chawla, Bowyer, Hall, and Kegelmeyer}]{smote}
Nitesh~V. Chawla, Kevin~W. Bowyer, Lawrence~O. Hall, and W.~Philip Kegelmeyer.
  2002.
\newblock {SMOTE:} synthetic minority over-sampling technique.
\newblock \emph{J. Artif. Intell. Res.}, 16:321--357.

\bibitem[{Dinkov et~al.(2019)Dinkov, Ali, Koychev, and
  Nakov}]{youtube-acoustic}
Yoan Dinkov, Ahmed Ali, Ivan Koychev, and Preslav Nakov. 2019.
\newblock Predicting the leading political ideology of {YouTube} channels using
  acoustic, textual, and metadata information.
\newblock In \emph{Proceedings of the 20th Annual Conference of the
  International Speech Communication Association}, INTERSPEECH~'19, pages
  501--505, Graz, Austria.

\bibitem[{Downs(1972)}]{Downs1972}
Anthony Downs. 1972.
\newblock Up and down with ecology: The ``issue-attention cycle''.
\newblock \emph{The Public Interest}, 28:38--50.

\bibitem[{Giachanou et~al.(2019)Giachanou, Rosso, and Crestani}]{Crestani:2012}
Anastasia Giachanou, Paolo Rosso, and Fabio Crestani. 2019.
\newblock Leveraging emotional signals for credibility detection.
\newblock In \emph{Proceedings of the 42nd International ACM SIGIR Conference
  on Research and Development in Information Retrieval}, SIGIR~'19, pages
  877--880, Paris, France.

\bibitem[{Gorrell et~al.(2018)Gorrell, Roberts, Greenwood, Bakir, Iavarone, and
  Bontcheva}]{10.1007/978-3-030-01129-1_17}
Genevieve Gorrell, Ian Roberts, Mark~A. Greenwood, Mehmet~E. Bakir, Benedetta
  Iavarone, and Kalina Bontcheva. 2018.
\newblock Quantifying media influence and partisan attention on {T}witter
  during the {UK EU} referendum.
\newblock In \emph{Social Informatics}, pages 274--290, Cham.

\bibitem[{Grinberg et~al.(2019)Grinberg, Joseph, Friedland, Swire-Thompson, and
  Lazer}]{Grinberg374}
Nir Grinberg, Kenneth Joseph, Lisa Friedland, Briony Swire-Thompson, and David
  Lazer. 2019.
\newblock Fake news on {T}witter during the 2016 {U.S.} presidential election.
\newblock \emph{Science}, 363(6425):374--378.

\bibitem[{Hoang et~al.(2011)Hoang, Lim, Achananuparp, Jiang, and
  Zhu}]{twitter-virality-modelling}
Tuan-Anh Hoang, Ee-Peng Lim, Palakorn Achananuparp, Jing Jiang, and Feida Zhu.
  2011.
\newblock On modeling virality of {T}witter content.
\newblock In \emph{Digital Libraries: For Cultural Heritage, Knowledge
  Dissemination, and Future Creation}, pages 212--221. Springer Berlin
  Heidelberg.

\bibitem[{Horne and Adalı(2017)}]{horne_adali_2017}
Benjamin~D. Horne and Sibel Adalı. 2017.
\newblock This just in: Fake news packs a lot in title, uses simpler,
  repetitive content in text body, more similar to satire than real news.
\newblock \emph{ArXiv 1703.09398}.

\bibitem[{Leskovec et~al.(2009)Leskovec, Backstrom, and
  Kleinberg}]{Leskovec2009}
Jure Leskovec, Lars Backstrom, and Jon Kleinberg. 2009.
\newblock Meme-tracking and the dynamics of the news cycle.
\newblock In \emph{Proceedings of the 15th ACM SIGKDD International Conference
  on Knowledge Discovery and Data Mining}, KDD~'09, pages 497--506, Paris,
  France.

\bibitem[{McLuhan(1964)}]{mcluhan1964understanding}
Marshall McLuhan. 1964.
\newblock \emph{Understanding media: The extensions of man}.
\newblock McGraw-Hill, New York.

\bibitem[{Munn(2019)}]{Munn2019}
Luke Munn. 2019.
\newblock Alt-right pipeline: Individual journeys to extremism online.
\newblock \emph{First Monday}.

\bibitem[{Nguyen et~al.(2020)Nguyen, Sugiyama, Nakov, and Kan}]{fang}
Van-Hoang Nguyen, Kazunari Sugiyama, Preslav Nakov, and Min-Yen Kan. 2020.
\newblock {FANG}: Leveraging social context for fake news detection using graph
  representation.
\newblock In \emph{Proceedings of the 29th ACM International Conference on
  Information and Knowledge Management}, CIKM~'20, pages 1165--1174.

\bibitem[{Reimers and Gurevych(2019)}]{sentence-bert}
Nils Reimers and Iryna Gurevych. 2019.
\newblock Sentence-{BERT}: Sentence embeddings using {S}iamese {BERT}-networks.
\newblock In \emph{Proceedings of the 2019 Conference on Empirical Methods in
  Natural Language Processing and the 9th International Joint Conference on
  Natural Language Processing}, EMNLP-IJCNLP~'19, pages 3982--3992, Hong Kong,
  China.

\bibitem[{Ribeiro et~al.(2020)Ribeiro, Ottoni, West, Almeida, and
  Meira~Jr}]{Ribeiro2020}
Manoel~Horta Ribeiro, Raphael Ottoni, Robert West, Virg{\'\i}lio~AF Almeida,
  and Wagner Meira~Jr. 2020.
\newblock Auditing radicalization pathways on {YouTube}.
\newblock In \emph{Proceedings of the 2020 Conference on Fairness,
  Accountability, and Transparency}, pages 131--141.

\bibitem[{Rubin et~al.(2016)Rubin, Conroy, Chen, and
  Cornwell}]{rubin_conroy_2016}
Victoria Rubin, Niall Conroy, Yimin Chen, and Sarah Cornwell. 2016.
\newblock Fake news or truth? {U}sing satirical cues to detect potentially
  misleading news.
\newblock In \emph{Proceedings of the Second Workshop on Computational
  Approaches to Deception Detection}, pages 7--17, San Diego, CA, USA.

\bibitem[{Ruchansky et~al.(2017)Ruchansky, Seo, and Liu}]{csi}
Natali Ruchansky, Sungyong Seo, and Yan Liu. 2017.
\newblock {CSI}: A hybrid deep model for fake news detection.
\newblock In \emph{Proceedings of the 2017 ACM on Conference on Information and
  Knowledge Management}, CIKM~'17, page 797–806, Singapore.

\bibitem[{Strubell et~al.(2019)Strubell, Ganesh, and
  McCallum}]{strubell-etal-2019-energy}
Emma Strubell, Ananya Ganesh, and Andrew McCallum. 2019.
\newblock Energy and policy considerations for deep learning in {NLP}.
\newblock In \emph{Proceedings of the 57th Annual Meeting of the Association
  for Computational Linguistics}, ACL~'19, pages 3645--3650, Florence, Italy.

\bibitem[{Venturini(2019)}]{Venturini2019}
Tommaso Venturini. 2019.
\newblock From fake to junk news, the data politics of online virality.
\newblock In Didier Bigo, Engin Isin, and Evelyn Ruppert, editors, \emph{Data
  Politics: Worlds, Subjects, Rights}. Routledge, London.

\bibitem[{Zhao et~al.(2015)Zhao, Resnick, and Mei}]{zhao_enquiring_2015}
Zhe Zhao, Paul Resnick, and Qiaozhu Mei. 2015.
\newblock Enquiring minds: Early detection of rumors in social media from
  enquiry posts.
\newblock In \emph{Proceedings of the 24th International Conference on World
  Wide Web}, WWW~'15, pages 1395--1405, Geneva, Switzerland.

\end{thebibliography}

\end{document}